\documentclass[sigconf,natbib=true]{acmart}
\usepackage{booktabs} % To thicken table lines
\usepackage[english]{babel}
\usepackage{moresize}
\usepackage{amsmath}
\usepackage{algorithmic}
\usepackage{balance}
\usepackage{comment}
\usepackage{paralist}
\usepackage{bm}
\usepackage{pgfplots}
\usetikzlibrary{pgfplots.dateplot}

\usepackage{flushend}
\usepackage[english]{babel}
\usepackage[latin1]{inputenc}
\usepackage{mathrsfs}
\usepackage{graphicx}

\usepackage{amssymb}
\usepackage{amsfonts}
\usepackage{url}
\usepackage{longtable}
\usepackage{rotating}
\usepackage{multirow}
\usepackage{mathrsfs}
\usepackage{subfigure}
\usepackage{enumitem}
\usepackage[linesnumbered,algoruled,boxed,lined]{algorithm2e}
\usepackage{adjustbox}
\usepackage{hyperref}
\usepackage{pgfplots}
\usetikzlibrary{pgfplots.dateplot}
\usepackage{filecontents}
\definecolor{tblue}{RGB}{31,119,180}
\definecolor{torange}{RGB}{255,127,14}
\definecolor{tgreen}{RGB}{44,160,44}
\definecolor{tred}{RGB}{214,39,40}
\definecolor{tpurple}{RGB}{148,103,189}

\newcommand{\hide}[1]{} %hide

\newcommand{\etal}{\textit{et al}.}

\newcommand{\ie}{\textit{i}.\textit{e}.}
\newcommand{\eg}{\textit{e}.\textit{g}.}

\usepackage{bbding}
\usepackage{threeparttable}
\usepackage{tabularx}
\usepackage{wrapfig}
\def\model{GraphGPT}

\graphicspath{{imag/}}

\setcopyright{none}
\copyrightyear{2024}
\acmYear{2024}
\setcopyright{acmlicensed}\acmConference[SIGIR'24]{Proceedings of the 47th
International ACM SIGIR Conference on Research and Development in
Information Retrieval}{July 14--18, 2024}{Washington, DC, USA}
\acmBooktitle{Proceedings of the 47th International ACM SIGIR Conference on
Research and Development in Information Retrieval (SIGIR'24), July 14--18,
2024, Washington, DC, USA}
\acmDOI{10.1145/3626772.3657775}
\acmISBN{979-8-4007-0431-4/24/07}

\begin{document}
\fancyhead{}




\title{GraphGPT: Graph Instruction Tuning for Large Language Models}
% \title{Graph Instruction Tuning for Large Language Models}

\author{Jiabin Tang}
\affiliation{%
  \institution{University of Hong Kong}
  \city{}
  \country{}}
\email{jiabintang77@gmail.com}

\author{Yuhao Yang}
\affiliation{%
  \institution{University of Hong Kong}
  \city{}
  \country{}}
\email{yuhao-yang@outlook.com}

\author{Wei Wei}
\affiliation{%
  \institution{University of Hong Kong}
  \city{}
  \country{}}
\email{weiwei1206cs@gmail.com}

\author{Lei Shi}
% \authornote{}
\affiliation{%
  \institution{Baidu Inc.}
  \city{}
  \country{}}
\email{harryshi.cs@gmail.com}

\author{Lixin Su}
% \authornote{}
\affiliation{%
  \institution{Baidu Inc.}
  \city{}
  \country{}}
\email{sulixinict@gmail.com}

\author{Suqi Cheng}
% \authornote{}
\affiliation{%
  \institution{Baidu Inc.}
  \city{}
  \country{}}
\email{chengsuqi@gmail.com}

\author{Dawei Yin}
% \authornote{}
\affiliation{%
  \institution{Baidu Inc.}
  \city{}
  \country{}}
\email{yindawei@acm.org}

\author{Chao Huang}
\authornote{Chao Huang is the Corresponding Author.}
\affiliation{%
  \institution{University of Hong Kong}
  \city{}
  \country{}}
\email{chaohuang75@gmail.com}

\renewcommand{\shorttitle}{GraphGPT: Graph Instruction Tuning for Large Language Models}
\renewcommand{\shortauthors}{Jiabin Tang et al.}

% \author{Anonymous Author(s)}
% \author{Jiabin Tang$^1$, Yuhao Yang$^2$, Wei Wei$^2$, Lei Shi$^3$, \\ Lixin Su$^3$, Suqi Cheng$^3$, Dawei Yin$^3$ and Chao Huang$^{1,2*}$}
% \affiliation{$^1$Musketeers Foundation Institute of Data Science, \\ $^2$Department of Computer Science,  University of Hong Kong, $^3$Baidu Inc.\\}
% \affiliation{\textbf{Project Page}: \href{https://GraphGPT.github.io}{https://GraphGPT.github.io}, \textbf{Github}: \href{https://github.com/HKUDS/GraphGPT}{https://github.com/HKUDS/GraphGPT}}

% \author{
%   Jiabin Tang\textsuperscript{$\clubsuit$}
% }


\begin{abstract}
Graph Neural Networks (GNNs) have evolved to understand graph structures through recursive exchanges and aggregations among nodes. To enhance robustness, self-supervised learning (SSL) has become a vital tool for data augmentation. Traditional methods often depend on fine-tuning with task-specific labels, limiting their effectiveness when labeled data is scarce. Our research tackles this by advancing graph model generalization in zero-shot learning environments. Inspired by the success of large language models (LLMs), we aim to create a graph-oriented LLM capable of exceptional generalization across various datasets and tasks without relying on downstream graph data. We introduce the GraphGPT framework, which integrates LLMs with graph structural knowledge through graph instruction tuning. This framework includes a text-graph grounding component to link textual and graph structures and a dual-stage instruction tuning approach with a lightweight graph-text alignment projector. These innovations allow LLMs to comprehend complex graph structures and enhance adaptability across diverse datasets and tasks. Our framework demonstrates superior generalization in both supervised and zero-shot graph learning tasks, surpassing existing benchmarks. The open-sourced model implementation of our \model\ is available at \color{blue}{\url{https://github.com/HKUDS/GraphGPT}}.
\end{abstract}



% \keywords{datasets, neural networks, gaze detection, text tagging}

\maketitle

\section{Introduction}
\label{sec:intro}

Graph neural networks (GNNs) have emerged as a powerful framework for analyzing and learning from graph-structured data~\cite{gnn_intro,gnn_intro_2}, enabling advancements in various domains, such as social network analysis~\cite{step_gnn,social_net}, recommender systems~\cite{lightgcn, gnn_rec_1}, and biological network analysis~\cite{drug_gnn, drug_gnn_2}. One of the key benefits of GNNs is their ability to capture the inherent structural information and dependencies present in graph data. By leveraging message passing and aggregation mechanisms, GNNs can effectively propagate and combine information across the graph, enabling them to model complex relationships and make accurate predictions.

In recent years, various GNN architectures have introduced innovations in how information is exchanged and aggregated among graph nodes. For example, graph convolutional network (GCNs)~\cite{gcn,gcn_intro_2} adapt convolutional operations to the graph domain, enabling effective feature representations. Graph attention networks (GATs)~\cite{gat,hgat} leverages attention mechanisms to assign different weights to neighboring nodes, allowing for more fine-grained information aggregation. Graph transformer networks (GTNs)~\cite{gt_1,HGT} incorporate self-attention and positional encoding to capture global dependencies and structural patterns in the graph. However, a notable limitation of many GNN approaches is their heavy reliance on supervised learning, which can lead to inadequate robustness and generalization when confronted with sparse and noisy data.

To enhance the generalization ability of GNNs, self-supervised learning (SSL) has emerged as a promising approach in graph representation learning. It aims to pre-train a robust graph model using auxiliary tasks on unlabeled graph data. The idea is to leverage the inherent structure and patterns within the graph itself to create meaningful self-supervisory signals. SSL-enhanced graph learning methods exhibit two primary paradigms: contrastive SSL and generative SSL. Within contrastive SSL, the emphasis lies on learning representations by contrasting positive and negative samples, with notable advancements of DGI~\cite{DGI} and GCA~\cite{zhu2021graph}. Conversely, generative SSL focuses on generating synthetic samples that closely resemble the original graph structures with masked autoencoders, exemplified by techniques like GraphMAE~\cite{hou2022graphmae} and S2GAE~\cite{tan2023s2gae}. 

While these methods aim to generate graph embeddings that are generalizable to different downstream tasks, they often require a fine-tuning process using labels specific to the downstream graph learning scenarios. However, this reliance on labeled data from downstream tasks can restrict their generalization in practical situations where obtaining high-quality labels may not always be feasible. This limitation is particularly relevant in learning scenarios like cold-start recommendation systems or traffic flow prediction in new cities where accurate labels may be scarce or unavailable.

As a result, the objective of this research is to advance the generalization capabilities of graph models by addressing challenging and practical zero-shot learning scenarios. Inspired by the remarkable success of large language models (LLMs) in natural language processing (NLP) tasks~\cite{LLMRec}, where they have demonstrated exceptional generalization abilities, this work aims to develop a graph-oriented LLM capable of achieving high generalization across diverse downstream datasets and tasks. However, effectively integrating large language models with graph learning poses non-trivial challenges. \vspace{-0.03in}

\begin{itemize}[leftmargin=*]

\item \textbf{C1}: Achieving a proper alignment between the structural information of a graph and the language space demands meticulous deliberation and thoughtful consideration. 

\item \textbf{C2}: Effectively guiding LLMs to comprehend the structural information of graphs remains a considerable challenge. 

\item \textbf{C3}: Endowing LLMs with the ability to reason step-by-step is important when tackling complex graph learning tasks.

\end{itemize}

\begin{figure}[t]
    \begin{center}
    %\framebox[4.0in]{$\;$}
    %\fbox{\rule[-.5cm]{0cm}{4cm} \rule[-.5cm]{4cm}{0cm}}
    \includegraphics[width=0.46\textwidth]{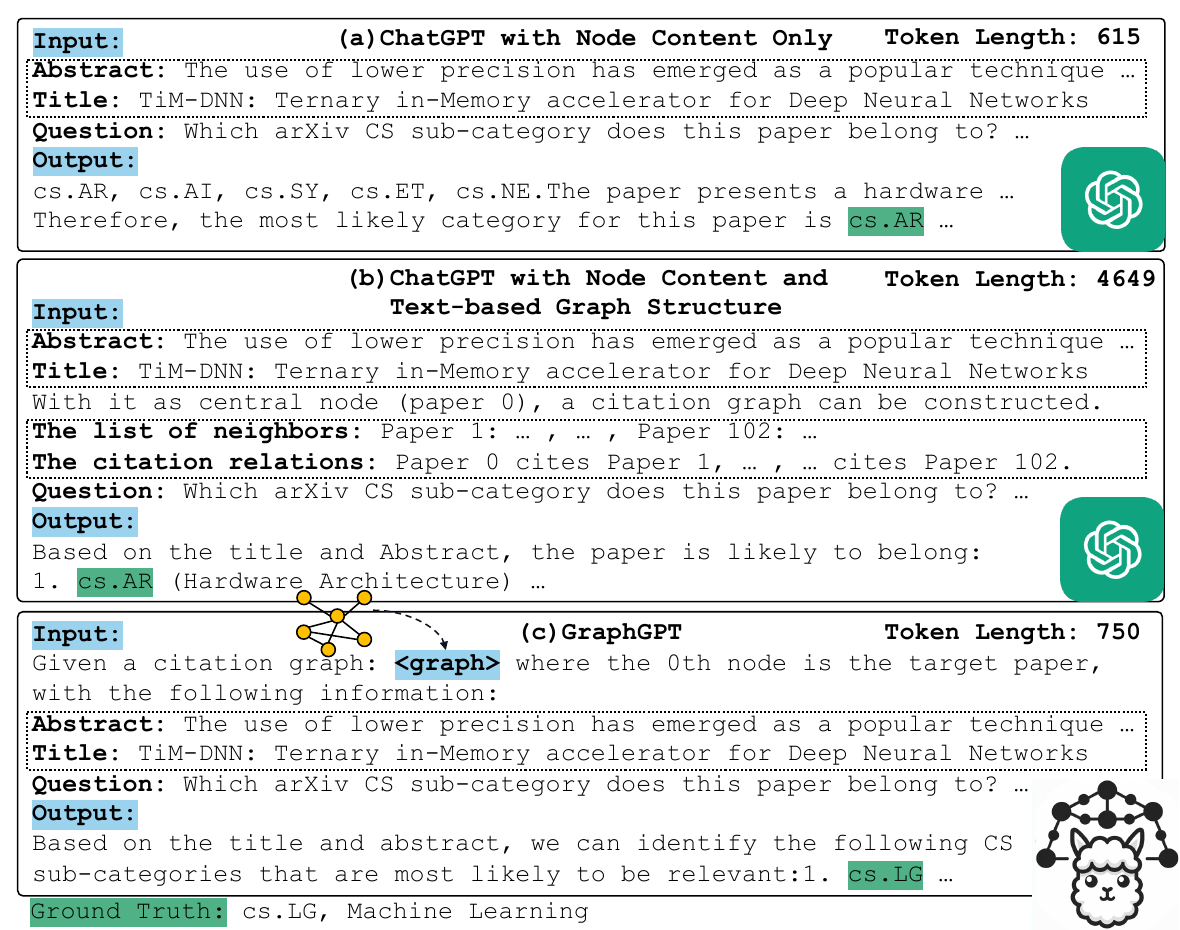}
    \end{center}
    \vspace{-3mm}
    \caption{Limitation of LLMs in understanding graph structural contexts with heavy reliance on textual data.}
    \label{fig:over}
    \vspace{-4.5mm}
\end{figure}

To gain a deeper understanding of the limitations associated with directly prompting LLMs using purely text-based prompts for graph structure modeling, we provide illustrative examples in Figure~\ref{fig:over}. These examples facilitate a comparative analysis between our \model\ framework and the ChatGPT approach. We focus on a representative node classification task, where the objective is to predict the category of a given paper. In Figure~\ref{fig:over} (a) and Figure~\ref{fig:over} (b), we showcase the prediction results for two scenarios using ChatGPT: (1) utilizing only the input node textual data, and (2) employing text-based graph structure-aware prompts inspired by the prompt designs in recent studies~\cite{GPT4Graph, potential_llm}. These figures highlight the potential limitations that arise when relying solely on text-based prompts for graph structure modeling, as evidenced by the incorrect paper node classification results presented. In contrast, our \model\ framework effectively addresses these limitations by preserving and leveraging the graph structural information, as shown in Figure~\ref{fig:over} (c). It enables accurate identification of the paper category, in understanding the underlying graph structure. 

Additionally, the utilization of text-based structural prompts leads to an increase in token size, which presents challenges in practical scenarios. Longer token sequences incur higher computational and memory costs, making it less feasible for real-world applications. Furthermore, existing LLMs have token limits, which further restrict the applicability of longer text-based prompts for large-scale graph structure modeling. These limitations emphasize the necessity for more efficient and scalable approaches that can effectively incorporate graph structural information into LLMs. \\\vspace{-0.12in}

\noindent \textbf{Contributions}. To address these challenges, we propose a novel framework called \model, which aims to align Large Language Models (LLMs) with Graphs using a carefully designed graph instruction tuning paradigm. (\textbf{C1}) Our framework introduces a text-graph grounding paradigm as the initial step to align the encoding of graph structures with the natural language space. By incorporating textual information in a contrastive manner, we enable effective alignment of graph structure information within language models. (\textbf{C2}) In our proposed dual-stage graph instruction tuning paradigm, we leverage self-supervised signals through the graph matching task, which is derived from unlabeled graph structures, to serve as instructions for guiding model tuning of LLMs. By incorporating this self-supervised instruction tuning, the language model acquires domain-specific structural knowledge related to graphs, thereby enhancing its understanding of graph structures. To further customize the LLM's reasoning behavior for diverse downstream graph learning tasks, the second stage of our graph instruction tuning paradigm involves fine-tuning the LLM with task-specific graph instructions, to improve the model's adaptability. (\textbf{C3}) By incorporating the Chain-of-Thought (COT) distillation into our framework, \model\ enhances its step-by-step reasoning abilities and improves its performance in the face of distribution shift.

In summary, our work makes the following contributions:
\begin{itemize}[leftmargin=*]

\item This work aims to align graph domain-specific structural knowledge with the reasoning ability of Large Language Models (LLMs) to improve the generalization of graph learning. \\\vspace{-0.12in}

\item Our approach aims to align LLMs with Graphs through a graph instruction tuning paradigm. This paradigm incorporates self-supervised instruction tuning, enhancing the LLM's comprehension of graph structural knowledge and its reasoning capabilities. Additionally, we introduce task-specific instruction tuning to improve the model's adaptability across diverse graph tasks. \\\vspace{-0.12in}

\item We evaluate our proposed \model\ on supervised and zero-shot graph learning tasks. We conduct thorough analyses of its component-wise effects and generalization ability. By comparing it with state-of-the-art baselines, we demonstrate the superior generalization power of our approach across various settings. 

\end{itemize}

% To ensure result reproducibility, we make our model implementation available at: \href{https://anonymous.4open.science/r/GraphGPT}{https://anonymous.4open.science/r/GraphGPT}.
% To ensure result reproducibility, we make our model implementation available at: \href{https://anonymous.4open.science/r/GraphGPT}{https://anonymous.4open.science/r/GraphGPT}.

\section{Preliminaries}
\label{sec:model}

\noindent\textbf{Graph-structured Data}. represents information as entities (nodes) and the relationships (edges) between them. A graph is denoted as $\mathcal{G}(\mathcal{V}, \mathcal{E}, \mathbf{A}, \mathbf{X})$, comprising key components. The node set $\mathcal{V}$ represents the collection of nodes, with $|\mathcal{V}| = N$ indicating the total number of nodes. The edge set $\mathcal{E}$ characterizes the relationships between nodes. The adjacency matrix $\mathbf{A} \in \mathbb{R}^{N \times N}$ encodes the graph's topology, with each element $A_{i,j}$ indicating the presence or absence of an edge between nodes $i$ and $j$. The feature matrix $\mathbf{X}\in \mathbb{R}^{N \times F}$ contains attribute or feature information associated with each node, where $F$ represents the feature dimensionality. \\\vspace{-0.12in}
% represents information as a collection of entities (nodes) and the relationships (edges) between them. A graph is formally denoted as $\mathcal{G}(\mathcal{V}, \mathcal{E}, \mathbf{A}, \mathbf{X})$, encompassing several core components. The node set $\mathcal{V}$ represents a collection of nodes, with $|\mathcal{V}| = N$ indicating the total number of nodes in the graph. The edge set $\mathcal{E}$ characterizes the relationships or connections between the nodes. The adjacency matrix $\mathbf{A} \in \mathbb{R}^{N \times N}$ encodes the topology of the graph, with each element $A_{i,j}$ indicating the presence or absence of an edge between nodes $i$ and $j$. The feature matrix $\mathbf{X}\in \mathbb{R}^{N \times F}$ contains the attribute or feature information associated with each node, where $F$ represents the dimensionality of features. \\\vspace{-0.12in}

% \noindent\textbf{Graph Neural Networks}. have emerged as a powerful framework for representation learning from graph-structured data. Unlike traditional neural networks that operate on grid-like data, GNNs can effectively capture and model the complex relationships and dependencies present in graphs. Specifically, GNNs leverage the inherent structure of the graph, consisting of nodes and edges, to learn expressive node representations through iterative message propagation and aggregation operations, presented as follows.
\noindent \textbf{Graph Neural Networks}. have become a powerful framework for learning representations from graph-structured data. Unlike traditional neural networks that process grid-like data, GNNs excel in capturing the intricate relationships and dependencies within graphs. They utilize the graph's structure-comprising nodes and edges-to derive expressive node representations through repeated message propagation and aggregation operations.
\begin{align}
    m_{v}^{(l)} &= \text{Propagate}^{(l)}(\{h_{u}^{(l -1)}: u \in \mathcal{N}(v)\}), \nonumber \\
    h_{v}^{(l)} &= \text{Aggregate}^{(l)} (h_{v}^{(l - 1)}, m_{v}^{(l)}) \label{eq:gnn}
\end{align}
% After applying the message passing and aggregation mechanism in Graph Neural Networks (GNNs), the encoded feature vector of node $v$ at the $l$-th layer is denoted as $h_{v}^{(l)}$. The $\text{Propagate}^{(l)}$ function performs message passing by aggregating information from the neighboring nodes of $v$ at the $l$-th layer. The $\text{Aggregate}^{(l)}$ function then combines the aggregated information with the previous layer's representation of node $v$ to generate the updated representation $h_{v}^{(l)}$. By encoding graph structural information with the learned representations, GNNs can be customized for various downstream graph learning tasks, such as node classification and link prediction. \\\vspace{-0.12in}
In Graph Neural Networks, the feature vector of node $v$ at layer $l$ is denoted as $h_{v}^{(l)}$. Message passing is performed by the $\text{Propagate}^{(l)}$ function, aggregating information from neighboring nodes of $v$ in layer $l$. The $\text{Aggregate}^{(l)}$ function combines this information with the previous layer's representation of node $v$ to update $h_{v}^{(l)}$. By incorporating graph structure into learned representations, GNNs can be tailored for tasks like node classification and link prediction. \\\vspace{-0.12in}

% \noindent\textbf{Pre-trained Graph Models}. with self-supervised learning harness the power of self-supervision to learn meaningful graph representations without relying on labeled data. The core idea is to design pretext tasks that generate additional supervision signals from the graph's intrinsic properties.
\section{Methodology}
\label{sec:solution}

\begin{figure*}
    \centering
    \includegraphics[width=0.95\linewidth]{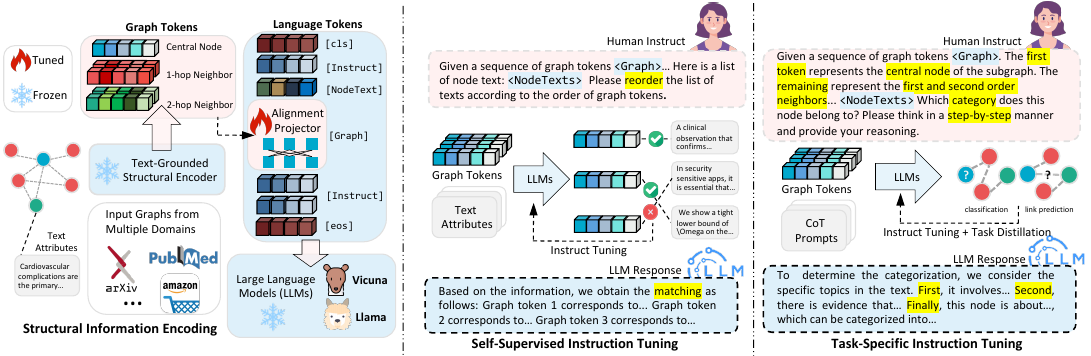}
    \vspace{-0.15in}
    \caption{The overall architecture of our proposed \model\ with graph instruction tuning paradigm.}
    \vspace{-0.1in}
    \label{fig:architecture}
\end{figure*}

\subsection{Structural Information Encoding with Text-Graph Grounding}
\label{sec:TG_ground}
To improve the understanding of graph structural information by large language models, our framework focuses on aligning the encoding of graph structures with the natural language space. This alignment enables language models to effectively comprehend the graph's structural elements using their language understanding capabilities. To achieve this, we introduce a text-graph grounding paradigm that generates prompts preserving the graph's structural context for language models. This paradigm acts as a bridge, connecting the semantic understanding of textual information with the inherent structural relationships in the graph.

In our \model, we design the graph encoder to be highly flexible, allowing it to leverage a wide range of backbone GNN architectures obtained from diverse graph pre-training paradigms. 
% To achieve an efficient text-graph grounding paradigm, we incorporate a non-parametric message-passing neural network as the structure-level pre-trained graph model, capturing the dependencies between nodes in the graph. In each message-passing step, the non-parametric graph encoder aggregates information from neighboring nodes, considering their relationships. The formal representation of the non-parametric message passing is as follows:
We incorporate a message-passing neural network architecture, which can be a graph transformer~\cite{gt_1} or a graph convolutional network~\cite{gcn}, as the structure-level pre-trained graph model. In each message-passing step, the graph encoder aggregates information from neighboring nodes, considering their relationships:
% \begin{align}
%     \mathbf{H}^{(l)} = \sigma\left( \tilde{\mathbf{D}}^{-\frac{1}{2}} \tilde{\mathbf{A}} \tilde{\mathbf{D}}^{-\frac{1}{2}} \mathbf{H}^{(l - 1)} \right)
%      \label{eq:mpnn}
% \end{align}
\begin{align}
    \mathbf{H}^{(l)} = \sigma\left( \tilde{\mathbf{A}} \mathbf{H}^{(l - 1)} \mathbf{W} \right)
     \label{eq:mpnn}
\end{align}
The self-loop adjacency matrix, denoted as $\tilde{\mathbf{A}}$, is obtained by adding the identity matrix $\mathbf{I}$ to the original adjacency matrix $\mathbf{A}$. $\mathbf{W}$ is the parameter matrix. This matrix captures the self-connections and local connectivity of nodes in the graph. $\sigma(\cdot)$ is the non-linear activation. $\mathbf{H}^{(l)}$ is the graph representations at the $l$-th layer. \\\vspace{-0.12in}

\noindent \textbf{Text-Structure Alignment}. 
% To align graph structure information within Language Models (LLMs) more effectively, our main focus is to explore the encoding of graph structures that can collaborate well with LLMs. Inspired by prior works~\cite{CLIP,G2P2}, we incorporate textual information into the graph structure encoding process in a contrastive manner. In our approach, we directly integrate a graph encoder with pre-trained parameters into our \model\ framework, enabling seamless integration of the graph encoder's capabilities in our framework. Formally, let $\mathcal{G}(\mathcal{V}, \mathcal{E}, \mathbf{A}, \mathbf{X})$ represent a graph with raw textual contents $\mathbf{C} = \{ c_i \in \mathbb{R}^{l_i\times d}, 1\leq i\leq N\}$ for $N$ nodes, where $l_i$ denotes the length of the textual content for the $i$-th node. We obtain encoded graph representations $\hat{\mathbf{H}}\in \mathbb{R}^{N\times d}$ and encoded text representations $\hat{\mathbf{T}} \in \mathbb{R}^{N\times d}$ as follows:
To enhance the alignment of graph structure information with Language Models (LLMs), our focus is on exploring effective encoding methods that can collaborate seamlessly with LLMs. Building upon previous works~\cite{CLIP, G2P2}, we adopt a contrastive approach by incorporating textual information into the graph structure encoding process. We directly integrate a pre-trained graph encoder into our \model\ framework, enabling the seamless utilization of its capabilities. Formally, given a graph $\mathcal{G}(\mathcal{V}, \mathcal{E}, \mathbf{A}, \mathbf{X})$ with raw textual contents $\mathbf{C} = { c_i \in \mathbb{R}^{l_i\times d}, 1\leq i\leq N}$ for $N$ nodes, we obtain encoded graph representations $\hat{\mathbf{H}}\in \mathbb{R}^{N\times d}$ and encoded text representations $\hat{\mathbf{T}} \in \mathbb{R}^{N\times d}$ as follows:
\begin{align}
    \mathbf{H} &= f_{\mathbf{G}} (\mathbf{X}), \mathbf{T} = f_{\mathbf{T}} (\mathbf{C}), \hat{\mathbf{H}} = \text{norm}(\mathbf{H}), \hat{\mathbf{T}} = \text{norm}(\mathbf{T}) 
    % \Gamma_i &= (g_i^{(1)}(\hat{\mathbf{H}}) g_i^{(2)}(\hat{\mathbf{T}})^{\top}) \cdot \exp (\tau) \nonumber \\
    % \mathcal{L} &= \sum_{i}\frac{1}{2} \lambda_i (\text{CE}(\Gamma_i, \mathbf{y}) + \text{CE}(\Gamma_i^{\top}, \mathbf{y}))
\end{align}
% We utilize the graph encoder, denoted as $f_{\mathbf{G}}$, to take the graph $\mathcal{G}(\mathcal{V}, \mathcal{E}, \mathbf{A}, \mathbf{X})$ as input and generates structure-level graph representations. We employ a text encoder, such as a transformer or Bert, denoted as $f_{\mathbf{T}}$, to encode the raw textual contents $\mathbf{C}$ associated with the nodes. This step produces encoded text representations of nodes. We further apply row-wise L2 normalization using the $\text{norm}$ function. Formally, the text-structure alignment with the cross-modalities is conducted as follows:
We utilize the graph encoder, $f_{\mathbf{G}}$, to generate structure-level graph representations from the input graph $\mathcal{G}(\mathcal{V}, \mathcal{E}, \mathbf{A}, \mathbf{X})$. To encode the raw textual contents $\mathbf{C}$ associated with the nodes, we employ a text encoder, such as a transformer or BERT, denoted as $f_{\mathbf{T}}$. This step produces encoded text representations of nodes, which are then normalized row-wise using the $\text{norm}$ function. The text-structure alignment across modalities is conducted as follows:
\begin{align}
    \Gamma_1 = (\hat{\mathbf{H}} \hat{\mathbf{T}}^{\top}) \cdot \exp (\tau)&, \Gamma_2 = (\hat{\mathbf{H}} \hat{\mathbf{T}}^{\prime\top}) \cdot \exp (\tau), \Gamma_3 = (\hat{\mathbf{T}}^{\top} \hat{\mathbf{T}}^{\prime\top}) \cdot \exp (\tau) \nonumber \\
    \mathcal{L} &= \sum_{i = 1}^{3}\frac{1}{2} \lambda_i (\text{CE}(\Gamma_i, \mathbf{y}) + \text{CE}(\Gamma_i^{\top}, \mathbf{y}))
\end{align}
% \begin{align}
%     \Gamma_1 &= (g_i^{(1)}(\hat{\mathbf{H}}) g_i^{(2)}(\hat{\mathbf{T}})^{\top}) \cdot \exp (\tau) \nonumber \\
%     \mathcal{L} &= \sum_{i}\frac{1}{2} \lambda_i (\text{CE}(\Gamma_i, \mathbf{y}) + \text{CE}(\Gamma_i^{\top}, \mathbf{y}))
% \end{align}
% Mathematically, these dimensions can be represented by three sets of transformation equations:
% \begin{align}
%     g_1^{(1)}(\hat{\mathbf{H}}) = \{\hat{\mathbf{H}}_i, 1\leq i\leq N\}&, g_1^{(2)}(\hat{\mathbf{T}}) =  \{\hat{\mathbf{T}}_i, 1\leq i\leq N\} \nonumber \\
%     g_2^{(1)}(\hat{\mathbf{H}}) = \{\hat{\mathbf{H}}_i, 1\leq i\leq N\}&, g_2^{(2)}(\hat{\mathbf{T}}) =  \{\frac{1}{|\mathcal{N}_i|}\sum_{j\in \mathcal{N}_i}\hat{\mathbf{T}}_j, 1\leq i\leq N\} \nonumber \\
%     g_3^{(1)}(\hat{\mathbf{H}}) = \{\hat{\mathbf{T}}_i, 1\leq i\leq N\}&, g_3^{(2)}(\hat{\mathbf{T}}) =  \{\frac{1}{|\mathcal{N}_i|}\sum_{j\in \mathcal{N}_i}\hat{\mathbf{T}}_j, 1\leq i\leq N\}
% \end{align}
% where $\hat{\mathbf{H}}\in \mathbb{R}^{N\times d}$ denotes graph representations and $\hat{\mathbf{T}} \in \mathbb{R}^{N\times d}$ indicates text representations, and $N$ is the number of nodes.
% In our text-graph grounding, we use the label $\mathbf{y} = (0, 1, \cdots, n-1)^{\top}$ for the contrastive alignment objective. The transformation functions for different grounding designs are denoted as $g_i^{(1)}$ and $g_i^{(2)}$. We employ a graph transformer~\cite{yun2019graph} as the graph encoder and a vanilla transformer~\cite{vaswani2017attention} as the text encoder.
where $\hat{\mathbf{T}}^{\prime} = \{\frac{1}{|\mathcal{N}_i|}\sum_{j\in \mathcal{N}_i}\hat{\mathbf{T}}_j, 1\leq i\leq N\}$ and $N$ is the number of nodes.
In our text-graph grounding, we use the label $\mathbf{y} = (0, 1, \cdots, n-1)^{\top}$ for the contrastive alignment objective. We employ a graph transformer~\cite{yun2019graph} as the graph encoder and a vanilla transformer~\cite{vaswani2017attention} as the text encoder.

% \begin{figure}[h]
%     \begin{center}
%     %\framebox[4.0in]{$\;$}
%     %\fbox{\rule[-.5cm]{0cm}{4cm} \rule[-.5cm]{4cm}{0cm}}
%     \includegraphics[width=0.32\textwidth]{imag/pre_ff_ok.pdf}
%     \end{center}
%     \vspace{-0.1in}
%     \caption{Workflow of text-structure alignment.}
%     \label{fig:align}
%     \vspace{-0.2in}
% \end{figure}

\vspace{-0.05in}
\subsection{Dual-Stage Graph Instruction Tuning}
% Recently, instruction tuning~\cite{Self_Instruct} is proposed to enhance the adaptability of language models for specific domains. Inspired by this, we design our dual-stage graph instruction tuning paradigm to further align the language capacity of the model with the nuances of graph learning task, allowing the language model to generate more accurate and contextually responses. Specifically, at the first stage, we introduce the self-supervised instruction tuning to facilitate the injection of graph domain-specific structural knowledge into the language model. By doing so, we ensure that the model has reasoning ability with the more graph contextually structural signals. Additionally, at the second stage, a task-specific instruction tuning is proposed customize the model's reasoning behavior to meet the specific constraints of specific tasks (\eg, node classification, link prediction). By fine-tuning LLM with our designed task-specific graph instructions, we can guide the model to generate responses that are better suited for the graph learning task at hand, which further improve the model adaptability to task-specific requirements.
The dual-stage graph instruction tuning paradigm proposed in this work builds upon the concept of instruction tuning, which has been recently introduced to enhance the adaptability of language models for specific domains~\cite{Self_Instruct}.  In this paradigm, we aim to align the language capacity of the model with the nuances of graph learning tasks, enabling the language model to generate more accurate and contextually appropriate responses for graph-structured data.

\vspace{-0.05in}
\subsubsection{\bf Self-Supervised Instruction Tuning}
% In the first stage of our graph instruction tuning paradigm, we introduce self-supervised instruction tuning. This mechanism allows us to inject graph domain-specific structural knowledge into the language model, improving its reasoning abilities and enabling it to effectively comprehend the contextual information embedded in the graph's structure. To achieve this, we incorporate self-supervised signals derived from unlabeled graph structures as instructions for model tuning. In particular, we design a structure-aware graph matching task that guides the language model in distinguishing between different graph tokens using natural language tokens. This instruction task plays a crucial role in accurately associating graph tokens with their corresponding textual descriptions, thereby deepening the model's understanding of the graph with the provided guidance. \\\vspace{-0.12in}
In the initial stage of our graph instruction tuning, we introduce self-supervised instruction tuning. This mechanism enhances the language model's reasoning abilities by incorporating graph domain-specific structural knowledge and effectively understanding contextual information within the graph's structure. To achieve this, we utilize self-supervised signals derived from unlabeled graph structures as instructions for model tuning. Specifically, we design a structure-aware graph matching task that guides the language model in differentiating between graph tokens using language tokens. This instruction task plays a vital role in accurately associating graph tokens with their corresponding textual descriptions, deepening the model's comprehension of the graph with the provided guidance. \\\vspace{-0.12in}

\noindent {\bf Instruction Design}. The instruction for our graph matching task consists of three components: i) graph information, ii) human question, and iii) \model\ response. In this task, we treat each node in the graph as a central node and perform h-hops with random neighbor sampling, resulting in a subgraph structure. The natural language input for the LLM is the human question. In the context of the graph matching task, the instruction includes the indicator token \texttt{<graph>} and a shuffled list of node text information. For example, in a citation graph, the node text information corresponds to paper titles. The objective of the LLM in the graph matching task is to align each graph token with its corresponding node text information. This requires reordering the node text information list based on the sequence of graph tokens, effectively associating each graph token with its relevant textual description. 
The detailed designs of graph matching are shown in Figure~\ref{fig:prompt}.
\\\vspace{-0.12in}

\noindent {\bf Tuning Strategy}.
To optimize the tuning process efficiently, we propose incorporating a \textbf{Lightweight Alignment Projector}. During training, we focus on optimizing the parameters of the projector $f_{\textbf{P}}$, while keeping the parameters of both the LLM and the graph encoder fixed. We assume that the projector successfully learns to map the encoded graph representation to graph tokens, while the LLM excels at aligning these tokens with diverse node text information. To align the graph tokens with the language tokens, we employ a projector $f_{\textbf{P}}$, which can be as simple as a single linear layer. This projector establishes correspondence between the graph tokens and the language tokens. By replacing the indicator token \texttt{<graph>} in the original language token sequence, the aligned graph tokens create a modified token sequence for the LLM. This modified sequence, denoted as $\{\texttt{<graph\_begin>}, \texttt{<graph\_token>}_1, \cdots, \texttt{<graph\_token>}_n, \\ \texttt{<graph\_end>}\}$, corresponds to the number of nodes $n$ in the graph associated with the given prompt. Given that the graph matching process is unsupervised, we have the opportunity to leverage a vast amount of unlabeled graph data from different domains, to enhance the generalizability of the learned projector. Mathematically, with projected graph tokens $\mathbf{X}_{\mathcal{G}} = f_{\mathbf{P}}(\mathbf{\hat{H}})$ and text embeddings $\mathbf{X}_{\mathcal{I}} = \text{tokenizer}(\text{instruction})$, for a sequence of length $L$, we compute the probability of generating the target output $\mathbf{X}_{\mathcal{O}}$ as follows:
% To optimize the tuning process efficiently, we propose incorporating a \textbf{Lightweight Alignment Projector}. During training, we focus on optimizing the parameters of the projector $f_{\textbf{P}}$, while keeping the parameters of both the LLM and the graph encoder fixed. We assume that the projector successfully learns to map the encoded graph representation to graph tokens, while the LLM excels at aligning these tokens with diverse node text information. By employing a simple projector like a single linear layer, we establish correspondence between the graph tokens and the language tokens to align them. The aligned graph tokens replace the indicator token $\texttt{<graph>}$ in the original language token sequence, resulting in a modified token sequence for the LLM. This modified sequence, denoted as $\texttt{<graph_begin>}$, 

% $\texttt{<graph_token>}_1$, $\cdots$, $\texttt{<graph_token>}_n, \texttt{<graph_end>}$, 

% corresponds to the number of nodes $n$ in the graph associated with the prompt. 

% The unsupervised nature of the graph matching process allows us to leverage a vast amount of unlabeled graph data from different domains, enhancing the generalizability of the learned projector. 

% Mathematically, with projected graph tokens $\mathbf{X}_{\mathcal{G}} = f_{\mathbf{P}}(\mathbf{\hat{H}})$ and text embeddings $\mathbf{X}_{\mathcal{I}} = \text{tokenizer}(\text{instruction})$ for a sequence of length $L$, we compute the probability of generating the target output $\mathbf{X}_{\mathcal{O}}$ as follows:
\begin{align}
    p(\mathbf{X}_{\mathcal{O}}|\mathbf{X}_{\mathcal{G}}, \mathbf{X}_{\mathcal{I}}) = \prod_{i=1}^{L}p_{\theta}(x_i|\mathbf{X}_{\mathcal{G}}, \mathbf{X}_{\mathcal{I}, <i} , \mathbf{X}_{\mathcal{O}, <i})
     \label{eq:lm_loss}
\end{align}
where $\theta$ are the learnable parameters within \model.

% In pursuit of a efficient tuning process, we propose our tuning strategy with a \textbf{lightweight alignment projector}. During the training process, we keep the parameters of both the language model (LLM) and the graph encoder fixed while optimizing only the parameters of the projector $f_{\textbf{G}}$. Once the training is completed, we assume that the projector has learned to accurately map the encoded graph representation to graph tokens, and the LLM is proficient in aligning these tokens with different node text information. To align the graph tokens with the natural language (NL) tokens, we use a projector $f_{\textbf{P}}$, which can be as simple as a single linear layer. The purpose of this projector is to establish the correspondence between the graph tokens and the NL tokens. The aligned graph tokens replace the indicator token \texttt{<graph>} in the original NL token sequence. This replacement results in a modified token sequence for the LLM, represented as 
% $\{\texttt{<graph\_begin>}, \texttt{<graph\_token>}_1, \cdots, \texttt{<graph\_token>}_n, \texttt{<graph\_end>}\}$, where $n$ denotes the number of nodes in the graph corresponding to the given prompt. Given that the graph match process is unsupervised, we have the opportunity to leverage a vast amount of unlabeled graph data from different domains, to enhance the generalizability of the learned projector.

\vspace{-0.05in}
\subsubsection{\bf Task-Specific Instruction Tuning}
% In the second stage, we propose task-specific instruction tuning. This step is designed to customize the model's reasoning behavior to meet the specific constraints and requirements of different graph learning tasks, such as node classification or link prediction. By fine-tuning the LLM using task-specific graph instructions, we guide the model to generate responses that are better suited for the particular graph learning task at hand. This further improves the model's adaptability and performance in handling various graph learning tasks. \\\vspace{-0.12in}
In the second stage, we introduce task-specific instruction tuning to customize the model's reasoning behavior for different graph learning tasks, such as node classification or link prediction. By fine-tuning the LLM using task-specific graph instructions, we guide the model to generate responses that align with the constraints and requirements of the specific graph learning task. This enhances the model's adaptability and performance in handling diverse graph learning tasks. \\\vspace{-0.12in}

\begin{figure}[h]
    \begin{center}
    %\framebox[4.0in]{$\;$}
    %\fbox{\rule[-.5cm]{0cm}{4cm} \rule[-.5cm]{4cm}{0cm}}
    \includegraphics[width=0.32\textwidth]{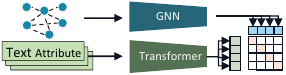}
    \end{center}
    \vspace{-0.1in}
    \caption{Workflow of text-structure alignment.}
    \label{fig:align}
    \vspace{-0.2in}
\end{figure}

\begin{figure*}[h]
    \begin{center}
    %\framebox[4.0in]{$\;$}
    %\fbox{\rule[-.5cm]{0cm}{4cm} \rule[-.5cm]{4cm}{0cm}}
    \includegraphics[width=0.98\textwidth]{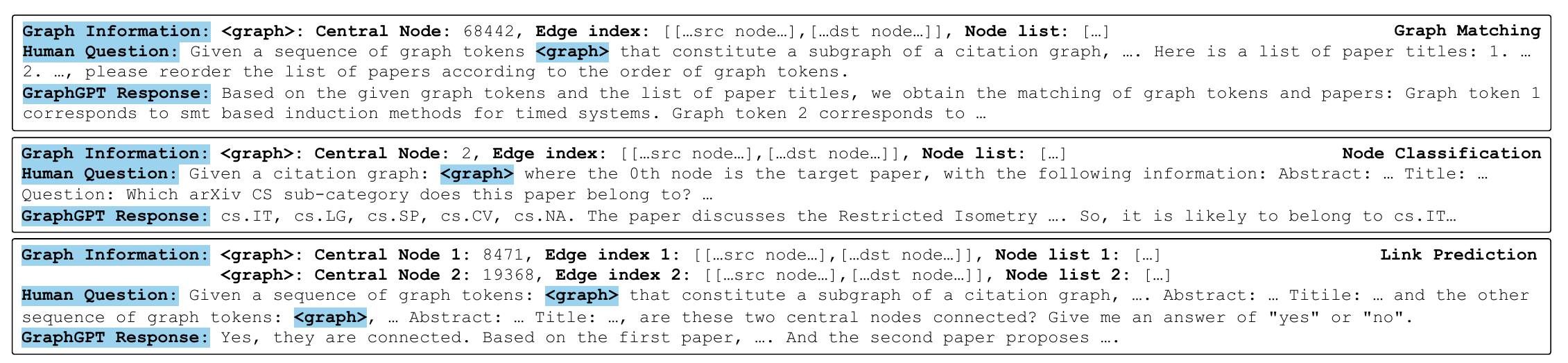}
    \end{center}
    \vspace{-0.1in}
    \caption{Our instruction designs for graph matching task (upper), node classification (middle) and link prediction (lower).}
    \label{fig:prompt}
    \vspace{-0.1in}
\end{figure*}

\noindent {\bf Instruction Design}.
% We adopt a similar instruction template, which consists of three parts. To generate graph information for each node, we employ the same neighbor sampling approach used in the first stage. This approach ensures that relevant graph information is captured, with each node acting as the central node. For the node classification task, the human question instruction contains both the indicator token \texttt{<graph>} and specific text information about the central node. This instruction prompts the language model to predict the category of the central node based on both the graph structure data and the accompanying text information. An example of the instruction data for different tasks can be seen in Figure~\ref{fig:prompt}, providing a visual representation of how the instruction is structured and presented to the language model. \\\vspace{-0.12in}
We utilize a consistent instruction template comprising three parts. To generate graph information for each node, we employ the same neighbor sampling approach as in the first stage. This approach ensures the inclusion of relevant graph information, with each node serving as the central node. For the node classification task, the human question instruction includes the indicator token \texttt{<graph>} and specific text information about the central node. This instruction guides the language model to predict the category of the central node based on both the graph structure data and the accompanying text information. Figure~\ref{fig:prompt} provides instruction examples for different tasks, visually illustrating the presentation of the instruction to the language model. \\\vspace{-0.12in}

\noindent {\bf Tuning Strategy}.
In the second stage of training, we utilize the parameters of the structure-aware projector that were trained in the first stage as the initial state. This allows us to conduct instruction tuning specifically for downstream tasks. During this training process, we keep the parameters of the language model (LLM) and graph encoder fixed, focusing solely on optimizing the parameters of the projector from the previous stage. By doing so, we ensure that the LLM further aligns with the requirements of downstream tasks, enhancing its ability to comprehend and interpret graph structures.

After completing the two training stages as described above, we have confidence that our \model\ has acquired the capability to comprehend the given graph structure and perform downstream tasks on the provided graph. The training process involving instruction tuning and the freezing of specific model parameters has refined the model's understanding of graph structures, enabling it to effectively tackle various tasks associated with the given graph.

\vspace{-0.1in}
\subsection{Chain-of-Thought (CoT) Distillation}
% When faced with diverse graph data, language models may encounter new or unfamiliar patterns and structures. This distribution shift can pose challenges in generating accurate and coherent responses, especially when the number of node classes varies across different types of graph data. To address this challenge and boost accuracy in the presence of distribution shift, it is essential to equip our \model\ with step-by-step reasoning abilities. In this regard, we propose utilizing the Chain-of-Thought (COT) technique~\cite{cot}, which explicitly models the flow of thoughts and reasoning steps. By incorporating COT, our language model improves the coherence and consistency of generated text. It enables the model to follow a logical progression of ideas, enhancing its ability to understand and reason about the given graph data. 
When faced with diverse graph data, language models may encounter unfamiliar patterns and structures, leading to challenges in generating accurate and coherent responses. This is especially true when the number of node classes varies across different types of graph data, causing distribution shift. To address this challenge and enhance accuracy in the presence of distribution shift, it is crucial to equip our \model\ with step-by-step reasoning abilities. Thus, we propose incorporating the Chain-of-Thought (COT) technique~\cite{cot}, which explicitly models the flow of thoughts and reasoning steps. By leveraging COT, our language model improves the coherence and consistency of generated text, enabling it to follow a logical progression of ideas and enhance its understanding and reasoning capabilities for the given graph data.

Incorporating the Chain-of-Thought (COT) technique can be challenging due to the influence of model parameter scale~\cite{COT_distill}. To overcome this, we draw inspiration from previous research~\cite{COT_distill} and adopt a distillation approach. By extracting valuable knowledge from a closed-source, powerful language model like ChatGPT (with over 200 billion parameters), we can generate high-quality COT instructions and enhance our model's COT reasoning capabilities without increasing the parameter count. \\\vspace{-0.12in}
% However, incorporating the Chain-of-Thought (COT) technique can be challenging due to the influence of model parameter scale~\cite{COT_distill}. To overcome this, we draw inspiration from previous research~\cite{COT_distill} and employ a distillation approach to extract valuable knowledge from a closed-source, powerful language model like ChatGPT (with over 200 billion parameters). This enables us to generate high-quality COT instructions and enhance our model's COT reasoning capabilities while avoiding an increase in parameters. \\\vspace{-0.12in}

% To address this challenge, we draw inspiration from~\cite{COT_distill} and involves leveraging closed-source, very powerful language models (LLMs) such as ChatGPT, which have over 200 billion parameters, to generate COT instruction data. These models have been trained on vast and diverse datasets, enabling them to provide high-quality COT instructions. However, instead of directly employing these large-scale models, we adopt a distillation approach.

\noindent \textbf{COT Distillation Paradigm}. 
% Our approach involves designing tailored Chain-of-Thought (COT) prompts specifically for node-specific tasks. For the node classification task within a citation graph, we provide the abstract, title of the paper represented by the node, and a description of the classification task as part of the input. We then employ the GPT-3.5 language model (LLM) in our implementation to perform step-by-step reasoning, by incorporating "\texttt{Please think about the categorization in a step by step manner.}" following the question. We prompt the LLM to arrive at the final answer by engaging in a sequential thought process. In the generated output, the LLM not only provides predictions for the node classes but also offers detailed explanations for each prediction. This ensures that the model's reasoning and decision-making process are transparent and comprehensible. To further improve the performance, we integrate the generated COT instruction data with the previously designed instructions for the task-specific instruction tuning stage. With the integrated instructions, we proceed with the proposed instruction tuning paradigm.
Our approach involves designing tailored Chain-of-Thought (COT) prompts for node-specific tasks. For the node classification task in a citation graph, we provide the abstract, paper title, and a task description as input. Using the GPT-3.5 language model (LLM), we incorporate "\texttt{Please think about the categorization in a step-by-step manner.}" to enable step-by-step reasoning. By engaging in sequential thought, the LLM generates output that includes predictions for node classes and detailed explanations for each prediction. This ensures transparent and comprehensible reasoning and decision-making. To further enhance performance, we integrate the generated COT instruction data with previously designed instructions for task-specific instruction tuning. With the integrated instructions, we proceed with the proposed instruction tuning paradigm.
% This information is presented in Table~\ref{tab:cot_prompt}. 
% We then employ the GPT-3.5 language model (LLM) in our implementation to perform step-by-step reasoning. We prompt the LLM to arrive at the final answer by engaging in a sequential thought process. In the generated output, the LLM not only provides predictions for the node classes but also offers detailed explanations for each prediction. This ensures that the model's reasoning and decision-making process are transparent and comprehensible. To further improve the performance, we integrate the generated COT instruction data with the previously designed instructions for the task-specific instruction tuning stage. With the integrated instructions, we proceed with the proposed instruction tuning paradigm.

% \input{solution_back}
% \vspace{-0.1in}
\vspace{-0.05in}   
\section{Evaluation}
\label{sec:eval}

% We conduct experiments to validate the effectiveness of our framework in various settings and address key research questions.
We conduct experiments to address key research questions:
\begin{itemize}[leftmargin=*]
\item \textbf{RQ1:} How does the proposed \model\ framework perform in both supervised and zero-shot graph learning settings?
\item \textbf{RQ2:} What is the generalization ability of our model in handling multiple tasks without experiencing catastrophic forgetting?
\item \textbf{RQ3:} What is the contribution of various key components in the proposed \model\ framework to its overall performance?
\item \textbf{RQ4:} How scalable and efficient is our \model\ framework?
\end{itemize}

\vspace{-0.05in}
\subsection{Experimental Settings}  
\subsubsection{\bf Data Descriptions}
% We evaluate the performance of our \model\ using three datasets: OGB-arxiv, PubMed, and Cora. The OGB-arxiv dataset~\cite{ogb} represents a directed graph that captures the citation network among computer science arXiv papers indexed by MAG~\cite{MAG}. Each paper in the dataset is associated with a research category, manually labeled by the authors and arXiv moderators. These research categories are selected from a set of 40 subject areas. The PubMed dataset~\cite{exp_as_feat} consists of 19,717 scientific publications on diabetes obtained from the PubMed database. The publications are categorized into Experimental induced diabetes, Type 1 diabetes, and Type 2 diabetes. Additionally, the dataset includes a citation network with 44,338 links. The Cora dataset~\cite{G2P2} comprises 25,120 research papers connected through citations. We utilize an expanded version of the Cora dataset, which is larger and has more classes (70 in total) compared to previous versions~\cite{gcn}.
We evaluate our \model\ using three datasets: OGB-arxiv, PubMed, and Cora. The OGB-arxiv dataset~\cite{ogb} represents a directed graph capturing the citation network among computer science arXiv papers indexed by MAG~\cite{MAG}. Each paper is manually labeled with a research category selected from 40 subject areas. The PubMed dataset~\cite{exp_as_feat} consists of 19,717 scientific publications on diabetes from the PubMed database, categorized into Experimental induced diabetes, Type 1 diabetes, and Type 2 diabetes. Additionally, it includes a citation network with 44,338 links. The Cora dataset~\cite{G2P2} comprises 25,120 research papers connected through citations. We use an expanded version with 70 classes, larger than previous versions~\cite{gcn}.

\vspace{-0.05in}   
\subsubsection{\bf Evaluation Protocols} 
% To ensure compatibility and enable comparison across different datasets, we map the node features into the same vector space. We achieve this by encoding the raw text information using a pre-trained BERT model~\cite{bert}. In our experiments, we divide the Cora and PubMed datasets into three parts: training, validation, and testing, following a ratio of 3:1:1. This partitioning scheme aligns with the approaches described in the works~\cite{exp_as_feat, G2P2}. For the OGB-arxiv dataset, we adhere to the public split setting~\cite{ogb}, which employs a ratio of 6:2:3 for training, validation, and testing. To evaluate the performance of our model, we utilize three commonly adopted evaluation metrics: Accuracy and Macro F1 for node classification, and AUC (Area Under the Curve) for link prediction.
To facilitate comparison across different datasets, we map node features into a unified vector space by encoding raw text information with a pre-trained BERT~\cite{bert}. In our experiments, we partition the Cora and PubMed datasets into training, validation, and testing sets following a 3:1:1 ratio, as described in previous works~\cite{exp_as_feat, G2P2}. For the OGB-arxiv data, we adhere to the public split setting~\cite{ogb} with a training, validation, and testing ratio of 6:2:3. To evaluate our model's performance, we utilize three commonly used metrics: Accuracy and Macro F1 for node classification, and AUC for link prediction.

\begin{table*}
    \centering
    \caption{Performance comparison of various methods on node classification under both supervised and zero-shot settings.}\label{tab:performance}
    \vspace{-0.1in}
    \resizebox{1\textwidth}{!}{\begin{tabular}{c|cc|cc|cc|cc|cc} 
    \hline
    Dataset                          & \multicolumn{2}{c|}{Arxiv-Arxiv}  & \multicolumn{2}{c|}{Arxiv-PubMed} & \multicolumn{2}{c|}{Arxiv-Cora}   & \multicolumn{2}{c|}{(Arxiv+PubMed)-Cora}  & \multicolumn{2}{c}{(Arxiv+PubMed)-Arxiv}\\ 
    \hline
    Model                            & Accuracy             & Macro-F1        & acc             & Macro-F1        & Accuracy             & Macro-F1        & Accuracy             & Macro-F1               & Accuracy             & Macro-F1  \\ 
    \hline
    MLP                              & 0.5179          & 0.2536          & 0.3940          & 0.1885          & 0.0258          & 0.0037          & 0.0220          & 0.0006                & 0.2127          & 0.0145   \\
    GraphSAGE                        & 0.5480          & 0.3290          & 0.3950          & 0.1939          & 0.0328          & 0.0132          & 0.0132          & 0.0029                & 0.1281          & 0.0129   \\
    GCN                              & 0.5267          & 0.3202          & 0.3940          & 0.1884          & 0.0214          & 0.0088          & 0.0187          & 0.0032                & 0.0122          & 0.0008   \\
    GAT                              & 0.5332          & 0.3118          & 0.3940          & 0.1884          & 0.0167          & 0.0110          & 0.0161          & 0.0057                & 0.1707          & 0.0285   \\
    RevGNN                            & 0.5474          & 0.3240          & 0.4440          & 0.3046          & 0.0272          & 0.0101          & 0.0217          & 0.0016                & 0.1309          & 0.0126   \\
    DGI                              & 0.5059           & 0.2787           & 0.3991           & 0.1905            & 0.0205            &  0.0011          &  0.0205           & 0.0011                 & 0.5059            & 0.2787   \\
    GKD                              & 0.5570          & 0.1595          & 0.3645          & 0.2561          & 0.0470          & 0.0093          & 0.0406          & 0.0037                 & 0.2089          & 0.0179  \\
    GLNN                             & 0.6088          & 0.3757          & 0.4298          & 0.3182          & 0.0267          & 0.0115          & 0.0182          & 0.0092                 & 0.3373          & 0.1115  \\
    NodeFormer                       & 0.5922          & 0.3328          & 0.2064          & 0.1678          & 0.0152          & 0.0065          & 0.0144          & 0.0053                 & 0.2713          & 0.0855  \\
    DIFFormer                        & 0.5986          & 0.3355          & 0.2959          & 0.2503          & 0.0161          & 0.0094          & 0.0100          & 0.0007                 & 0.1637          & 0.0234  \\
    baichuan-7B                      & 0.0946          & 0.0363          & 0.4642          & 0.3876          & 0.0405          & 0.0469          & 0.0405          & 0.0469                 & 0.0946          & 0.0363  \\
    vicuna-7B-v1.1                   & 0.2657          & 0.1375          & 0.5251          & 0.4831          & 0.1090          & 0.0970          & 0.1090          & 0.0970                 & 0.2657          & 0.1375  \\
    vicuna-7B-v1.5                   & 0.4962          & 0.1853          & 0.6351          & 0.5231          & 0.1489          & 0.1213          & 0.1489          & 0.1213                 & 0.4962          & 0.1853  \\ 
    \hline
    \textbf{\model-7B-v1.1-cot}        & 0.4913          & 0.1728          & 0.6103          & 0.5982          & 0.1145          & 0.1016          & 0.1250          & 0.0962                 &  0.4853       & 0.2102    \\
    \textbf{\model-7B-v1.5-stage2} & \textbf{0.7511} & \textbf{0.5600} & 0.6484          & 0.5634          & 0.0813          & 0.0713          & 0.0934          & 0.0978                 & 0.6278        &  0.2538    \\
    \textbf{\model-7B-v1.5-std}  & 0.6258          & 0.2622          & \textbf{0.7011} & \textbf{0.6491} & 0.1256          & 0.0819          & 0.1501          & 0.0936                 & 0.6390          & 0.2652      \\
    \textbf{\model-7B-v1.5-cot}    & 0.5759          & 0.2276          & 0.5213          & 0.4816          & \textbf{0.1813} & \textbf{0.1272} & \textbf{0.1647} & \textbf{0.1326}         &  \textbf{0.6476}   &  \textbf{0.2854}        \\ 
    \hline
    p-val                     &  2.26$e^{-9}$         &  1.56$e^{-10}$       &  2.22$e^{-7}$      &  1.55$e^{-9}$        &    1.04$e^{-9}$      &   9.96$e^{-6}$      &  7.62$e^{-8}$      &   1.97$e^{-7}$           &    1.5e$^{-13}$    &  4.63$e^{-6}$     \\
    \hline
     
    \end{tabular}
    }
    % \vspace{-0.1in}
    \end{table*}

\subsubsection{\bf Baseline Methods}
In our performance comparison, we consider various state-of-the-art methods for comprehensive evaluation. (i) The first category includes MLP, which employs a Multilayer Perceptron for node representation. (ii) The second category comprises representative graph neural encoders, including GraphSAGE~\cite{graphsage}, GCN~\cite{gcn}, GAT~\cite{gat}, and RevGNN~\cite{revgnn}. (iii) The third category focuses on the self-supervised approach DGI~\cite{DGI} for graph learning. (iv) The fourth category explores knowledge distillation-enhanced GNNs, with GKD~\cite{GeoKD} and GLNN~\cite{GLNN} as notable methods. (v). The fifth category showcases recently proposed strong graph transformer networks, with NodeFormer~\cite{NodeFormer} and DIFFormer~\cite{DIFFormer} as competitors. (vi) Lastly, we consider open-sourced LLMs, such as Baichuan-7B, vicuna-7B-v1.1, and vicuna-7B-v1.5 as baselines for understanding text-attributed graph data. 
% For more detailed descriptions of the baselines, please refer to the Appendix.

\vspace{-0.05in}
\subsubsection{\bf Implementation Details}
% For our model implementation, we primarily utilize the PyTorch and Transformers libraries. We employ Vicuna-7B-v1.1 and Vicuna-7B-v1.5 as the base models for our approach. The batch size is set to 2 on each GPU, and the learning rate is $2e^{-3}$. We apply a warmup ratio of $3e^{-2}$ and set the maximum input length of the Large Language Model (LLM) to 2048. The training process is carried out for 3 epochs. In the stage of task-specific instruction tuning, we explore the performance of the model under different data mixtures by adopting various combinations of instruction data. The hyperparameter settings remain constant, except for the number of training epochs, which is set to 2 in this stage. The alignment projector parameters fine-tuned in the self-supervised instruction tuning stage are used as the initial parameters for the projector in the second tuning stage. For the evaluation of most baselines, we utilize their publicly available code. We employ a grid-search strategy based on their default hyperparameter settings to ensure a comprehensive evaluation. For further implementation details, please refer to our released source code.
For our model implementation, we primarily use the PyTorch and Transformers libraries. We utilize Vicuna-7B-v1.1 and Vicuna-7B-v1.5 as the base models. The batch size is set to 2 per GPU, and the learning rate is $2e^{-3}$. We apply a warmup ratio of $3e^{-2}$ and set the maximum input length of the Large Language Model (LLM) to 2048. The training process runs for 3 epochs. In the task-specific instruction tuning stage, we explore various combinations of instruction data to assess the model's performance under different data mixtures. The hyperparameter settings remain constant, except for the number of training epochs, which is set to 2 in this stage. The alignment projector parameters fine-tuned in the self-supervised instruction tuning stage serve as the initial parameters for the projector in the second tuning stage. For evaluating most baselines, we use their publicly available code. For more implementation details, please refer to our released code.

\vspace{-0.15in}
\subsection{Overall Performance Comparison (RQ1)}
% We conduct experiments on the node classification task, evaluating both supervised and zero-shot scenarios. The overall performance is presented in Table~\ref{tab:performance}. \textbf{Supervised Task Settings}: We train the models on a specific dataset and evaluated their performance on the corresponding test set (\eg, training on Arxiv-Arxiv and testing on the Arxiv test set). \textbf{Zero-Shot Task Settings}: We train the models on a specific dataset and test them on other datasets without any additional training (\eg, training on Arxiv-PubMed and testing on the PubMed dataset). To account for variations in the number of classes across different datasets, we employed a classifier trained with transfer data, typically a linear layer, when testing GNN-based models. In Table~\ref{tab:performance}, "-7B-" represents the parameter scale, while "-v1.1-" and "-v1.5-" indicate different versions of the base Vicuna model. "-stage2" indicates that only the second stage tuning is adopted. "-std" and "-cot" denote the use of the standard and generated COT instruction datasets, respectively. \\\vspace{-0.1in}
We conduct experiments on the node classification task, evaluating both supervised and zero-shot scenarios. The overall performance is summarized in Table~\ref{tab:performance}. In the \textbf{Supervised Task Setting}, models are trained on a specific dataset and evaluated on the corresponding test set (e.g., training on Arxiv-Arxiv and testing on the Arxiv test set). In the \textbf{Zero-Shot Task Setting}, models are trained on a specific dataset and tested on other datasets without additional training (e.g., training on Arxiv-PubMed and testing on the PubMed dataset). To handle variations in the number of classes across datasets, we employ a transfer-trained classifier, such as a linear layer, when testing GNN-based models. In Table~\ref{tab:performance}, "-7B-" indicates the parameter scale, while "-v1.1-" and "-v1.5-" represent different versions of the base Vicuna model. "-stage2" indicates the adoption of only the second stage tuning. "-std" and "-cot" denote the use of the standard and generated COT instruction datasets, respectively. \\\vspace{-0.1in}

\noindent \noindent\textbf{Obs.1: Overall Superiority of our \model.} 
Our graph LLM consistently outperforms various state-of-the-art baselines in both supervised and zero-shot scenarios. Notably, even recently developed strong GNN-based models, such as NodeFormer, DIFFormer, and GKD, exhibit good structural modeling capabilities in the supervised setting. However, when transferred to new datasets without further training, their performance significantly declines. In contrast, our \model\ not only surpasses all state-of-the-art methods in supervised tasks but also achieves a remarkable 2-10 times increase in accuracy in the zero-shot graph learning scenario. 

% Additionally, LLM-based solutions, like Baichuan-7B and Vicuna-7B maintain stable performance across different datasets. However, they are limited to making predictions solely based on text information. In contrast, our \model\ effectively preserves graph structured information, offers a more comprehensive solution for graph learning tasks. These improvements can be attributed to two key factors: i) Through our dual-stage graph instruction tuning, our method aligns the graph tokens, which contain rich structural information encoded by the graph encoder, with the natural language tokens. This alignment allows the LLM to retain and understand the inherent structural characteristics of the graph data. ii) Our framework facilitates mutual enhancement between the graph encoder and LLM. The introduction of graph tokens fills the gap in the LLM's structural understanding, enabling it to incorporate and reason about the graph's structural information. \\\vspace{-0.1in}
LLM-based solutions like Baichuan-7B and Vicuna-7B maintain stable performance across different datasets but rely solely on text information for predictions. In contrast, our \model\ preserves graph structure, providing a comprehensive solution for graph learning tasks. Two key factors contribute to these improvements: (i) Our dual-stage graph instruction tuning aligns structural information encoded by the graph encoder with natural language tokens, enabling the LLM to understand the graph's inherent characteristics. (ii) Our framework facilitates mutual enhancement between the graph encoder and LLM, filling the LLM's gap in structural understanding and enabling it to reason about the graph's structure. \\\vspace{-0.1in}

\noindent\textbf{Obs.2: Benefits with Structure-aware Graph Matching.}
The presence of the first stage, which involves self-supervised graph matching tasks for instruction tuning, plays a crucial role in enhancing the zero-shot transferability of our \model. The first stage focuses on aligning the graph tokens, which encode rich structural information, with the language tokens. This alignment enables the model to develop a deeper understanding of the inherent structural characteristics of the graph data. Without the first stage, if we only conduct the second stage of task-specific instruction tuning, the model tends to be more prone to overfitting on the specific dataset. In such cases, the model's performance may be heavily reliant on dataset-specific patterns and characteristics, rather than a genuine understanding of the underlying graph structure. This can limit the model's ability to generalize to new, unseen datasets. \\\vspace{-0.1in}

\noindent\textbf{Obs.3: Benefits with COT Distillation.}
The "-std" and "-cot" variants indicate that the use of COT distillation substantially benefits more complex graph learning tasks. Models tuned with the standard instruction dataset can already achieve prominent results when transferred to simpler tasks, such as the PubMed dataset with 3 classes, with an accuracy of 0.7011 for Arxiv-PubMed. However, their performance tends to be mediocre when applied to complex tasks like the Cora dataset with 70 classes. By leveraging the powerful reasoning capabilities of the closed-source model (GPT-3.5) through COT distillation, our model can integrate this knowledge and significantly enhance its performance on complex graph tasks.

\begin{table}
\centering
\caption{Performance comparison of various instruction mixtures in supervised learning on the Arxiv dataset and the zero-shot setting on the Cora dataset for node classification.}\label{tab:mix_1}
\vspace{-0.1in}
\resizebox{0.40\textwidth}{!}{\begin{tabular}{c|cc|cc} 
\hline
Dataset                       & \multicolumn{2}{c|}{Supervision. on Arxiv}      & \multicolumn{2}{c}{Zero Shot on Cora}   \\ 
\hline
Model                         & Acc               & Macro-F1          & Acc               & Macro-F1           \\ 
\hline
MLP                           & 0.5179            & 0.2536            & 0.0220            & 0.0006             \\
GraphSAGE                     & 0.5480            & 0.3290            & 0.0132            & 0.0029             \\
GCN                           & 0.5267            & 0.3202            & 0.0187            & 0.0032             \\
GAT                           & 0.5332            & 0.3118            & 0.0161            & 0.0057             \\
RvGNN                         & 0.5474            & 0.3240            & 0.0217            & 0.0016             \\
DGI                           & 0.5059            & 0.2787            & 0.0205            & 0.0011             \\
GKD                           & 0.5570            & 0.1595            & 0.0406            & 0.0037             \\
GLNN                          & 0.6088            & 0.3757            & 0.0182            & 0.0092             \\
NodeFormer                    & 0.5922            & 0.3328            & 0.0144            & 0.0053             \\
DIFFormer                     & 0.5986            & 0.3355            & 0.0100            & 0.0007             \\
baichuan-7b                   & 0.0946            & 0.0363            & 0.0405            & 0.0469             \\
vicuna-7B-v1.1                & 0.2657            & 0.1375            & 0.1090            & 0.0970             \\
vicuna-7B-v1.5                & 0.4962            & 0.1853            & 0.1489            & 0.1213             \\ 
\hline
Arxiv-std + PubMed-std          & 0.6390            & 0.2652            & 0.1501            & 0.0936             \\
Arxiv-cot + PubMed-cot        &   0.6476           &  0.2854           & 0.1647            & 0.1326             \\
Arxiv-mix + PubMed-mix        & 0.6139            &  0.2772          & 0.1544          & 0.1048           \\
Arxiv-std + PubMed-std + Link & 0.5931          & 0.2238           & \textbf{0.1847} & \textbf{0.1579}  \\
Arxiv-mix + Pubmed-mix + Link & \textbf{0.6874} & \textbf{0.3761} & 0.1836           & 0.1494           \\
\hline
\end{tabular}}
\vspace{-0.15in}
\end{table}

\begin{table}
\centering
\caption{Performance comparison of various instruction mixtures for link prediction on PubMed.}\label{tab:mix_2}
\vspace{-0.1in}
\resizebox{0.35\textwidth}{!}{\begin{tabular}{c|cc} 
\hline
Dataset                       & \multicolumn{2}{c}{PubMed}  \\ 
\hline
Model                         & AUC    & AP                 \\ 
\hline
MLP                           & 0.5583 & 0.5833             \\
GAT                           & 0.5606 & 0.6373             \\
GraphSAGE                          & 0.5041 & 0.5813             \\
RevGNN                        & 0.4538 & 0.5083             \\
Node2Vec                      & 0.6535 & 0.6885             \\ 
\hline
w/o Link                       & 0.5010 & 0.5005              \\
only Link                     & 0.6704  &  0.6087            \\
Arxiv-std + PubMed-std + Link & \textbf{0.8246} & \textbf{0.8026}             \\
Arxiv-mix + PubMed-mix + Link & 0.6451 & 0.5886             \\
\hline
\end{tabular}}
\end{table}

\vspace{-0.05in}
\subsection{Generalization Ability Investigation (RQ2)}
In this subsection, we explore the generalization ability of our model by incorporating more instruction data to fine-tune the LLM for effectively handling various types of tasks. Our main results and experimental observations are presented as follows: \\\vspace{-0.12in}

\noindent\textbf{More Data Boost Model Transfer Ability.}
In our preliminary investigation, we examine the influence of data quantity on the transfer capability of our \model, as illustrated in the "(Arxiv + PubMed)-Cora" column of Table~\ref{tab:performance}. In this experiment, we train models using a combination of the Arxiv and PubMed datasets and perform zero-shot testing on the Cora dataset. The results reveal that by incorporating a relatively smaller PubMed dataset (with 20,000+ items) alongside Arxiv, our \model\ exhibits a significant improvement in transfer performance on Cora. In contrast, the transfer performance of GNN-based models, trained separately on Arxiv and PubMed, actually deteriorates. \\\vspace{-0.12in}

\noindent\textbf{More Data Yet No Forgetting.}
We further validate the performance of the combined Arxiv and PubMed instruction data on the original Arxiv data, as demonstrated in the "(Arxiv + PubMed)-Arxiv" column in Table~\ref{tab:performance}. The results indicate that most traditional GNN-based approaches experience a significant decline in performance on Arxiv after iterative training. In contrast, our model exhibits improved performance. We attribute this phenomenon to the occurrence of catastrophic forgetting in GNN-based models, where the structural modeling competence of the model trained solely on the smaller PubMed dataset is compromised. However, our model effectively mitigates this issue through our unified graph instruction tuning paradigm. This enables our model to maintain and even enhance its performance by retaining the generalized graph structure patterns despite incorporating additional data. \\\vspace{-0.12in}

\noindent\textbf{Generalization for Multitasking Graph Learner.} Recent studies on instruction tuning suggest that mixing different instruction tuning data can further enhance the performance of Language and Logic Models (LLMs). In this study, we ensure a consistent number of instruction entries and mix different types of instruction data, including standard instruction ("-std"), COT instruction ("-cot"), a blend of standard (50\%) and COT (50\%) instruction ("-mix"), and link prediction instruction ("Link"). The results are presented in Tables~\ref{tab:mix_1} and Table~\ref{tab:mix_2}. We can observe that effective data mixture solutions can significantly improve the performance of our \model\ under various settings. The addition of task-specific instruction for link prediction task notably enhances the performance of our model in node classification. Interestingly, after incorporating node classification, the performance of link prediction also exceeds that of the selected best-performing existing models. After mixing the instructions of different tasks, our model demonstrates the ability to effectively handle various graph learning tasks and transfer its knowledge to other unseen datasets.

\begin{table}
\centering
\caption{Module ablation study under both supervised and zero-shot settings to analyze the individual contributions.}\label{tab:ablation}
\footnotesize
\vspace{-0.1in}
\resizebox{.48\textwidth}{!}{\begin{tabular}{c|cc|cc|cc} 
\hline
Dataset            & \multicolumn{2}{c|}{Arxiv-Arxiv}  & \multicolumn{2}{c|}{Arxiv-PubMed}   & \multicolumn{2}{c}{Arxiv-Cora}   \\ 
\hline
Variant            & Acc             & Mac-F1        & Acc             & Mac-F1        & Acc             & Mac-F1         \\
\hline
w/o GS & 0.4962          & 0.1853          & 0.6351          & 0.5231          & 0.1489          & 0.1213           \\
% w mpnn               & 0.6064          & 0.2439          & 0.6159          & 0.4703          & 0.1029          & 0.0760           \\ 
w/o LR               & 0.5807          & 0.2462          & 0.2523          & 0.1925          & 0.0050          & 0.0016           \\
% mpnn only               & 0.3928          & 0.1533          & 0.3818          & 0.2801          & 0.0281          & 0.0044           \\
\hline
\textbf{ours}      & \textbf{0.6258} & \textbf{0.2622} & \textbf{0.7011} & \textbf{0.6491} & \textbf{0.1813} & \textbf{0.1272}  \\
\hline
\end{tabular}}
\vspace{-0.1in}
\end{table}
% \vspace{-0.05in}

\subsection{Module Ablation Study (RQ3)}
We conduct an ablation study to investigate the individual contributions of different sub-modules of our proposed framework, and the results are reported in Table~\ref{tab:ablation}. The observations are as follows: \\\vspace{-0.1in}

\noindent \textbf{Effect of Graph Instruction Tuning}. In our study, we investigate the benefit of incorporating graph structural information into LLM using the variant "w/o GS." In this variant, we directly adopt the base LLM (specifically, Vicuna-7B-v1.5) to perform node classification on three datasets, without incorporating graph structural information. The results of our study demonstrate that our model significantly outperforms the base model that lacks structural information. This indicates that our graph instruction tuning paradigm enables the LLM to understand the graph structural information more effectively. Importantly, this improvement in performance was achieved without altering the original parameters of the LLM. Instead, it was solely accomplished through our lightweight alignment projector, which aligns graph tokens and natural language tokens through the 1-linear projection operation. \\\vspace{-0.12in}

\noindent\textbf{Effect of LLM-enhanced Semantic Reasoning.}
We conduct further investigations to assess the influence of the LLM's reasoning ability in our \model\ by performing supervised and zero-shot predictions using only the default graph encoders. This variant is denoted as "w/o LR". The results of our study indicate that our \model, which integrates the LLM, significantly enhances the performance of graph encoder, especially in the zero-shot setting. This suggests that the rich semantic information injected by the LLM provides a substantial gain in performance.

\begin{table}
\centering
\caption{Study on the time and space efficiency of our \model\ during both the training and inference stages.}\label{tab:ablation}
\vspace{-0.1in}
\resizebox{.45\textwidth}{!}{\begin{tabular}{l|l|l|l} 
\hline
Variants            & Training Time & Tuned Parameters & GPU Occupy  \\ 
\hline
Stage-1-tune        & OOM           & 6,607,884,288    & OOM         \\
Stage-1-freeze      & 22:53:33      & 131,612,672      & 39517.75    \\ 
\hline
improvement         & ~-~           & $\downarrow\times$ 50.21          & ~ -~        \\ 
\hline
Stage-2-tune        & OOM           & 6,607,884,288    & OOM         \\
Stage-2-freeze      & 03:44:35      & 131,612,672      & 38961.75    \\ 
\hline
improvement         & ~-~           & $\downarrow\times$ 50.21          & ~-~         \\
\hline
\end{tabular}}
\vspace{-0.15in}
\end{table}

\vspace{-0.05in}
\subsection{Model Efficiency Study (RQ4)}
The study aims to assess the computational efficiency of our model during both the model training and inference stages. \\\vspace{-0.1in}

\noindent \textbf{Training Efficiency with Graph Instruction Tuning}. Our instruction tuning framework follows a two-stage process where the parameters of both the LLM and the graph encoder are frozen, and only the graph-text alignment projector is tuned. We conduct a comparison between freezing and tuning the LLM parameters in a 4-card 40G Nvidia A100 environment, denoted by "-freeze" and "-tune" respectively. The study analyze the time and space efficiency in terms of training time, the number of tuned parameters, and GPU occupancy (MiB per GPU). Under the same experimental conditions, when tuning LLM parameters, we encounter Out of Memory (OOM) errors even with a batch size of 1. However, by utilizing our tuning strategy, the training process remains stable even with a batch size of 2. Moreover, the number of tuned parameters decreases by more than 50 times compared to the freezing stage. \\\vspace{-0.1in}

\begin{figure}[t]
    \begin{center}
    %\framebox[4.0in]{$\;$}
    %\fbox{\rule[-.5cm]{0cm}{4cm} \rule[-.5cm]{4cm}{0cm}}
    \includegraphics[width=0.46\textwidth]{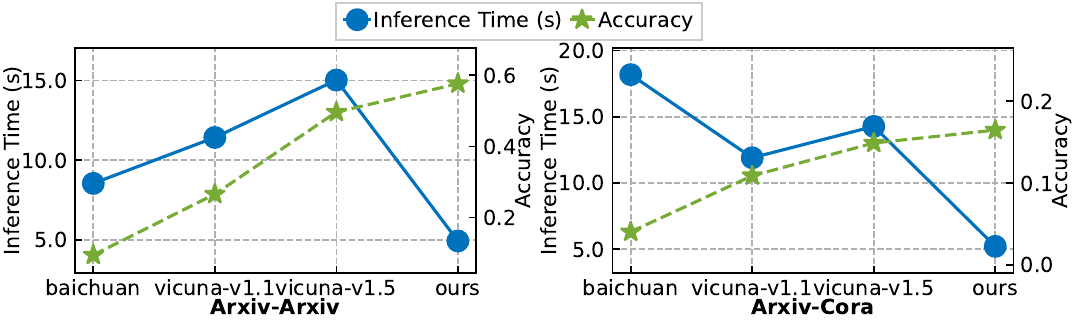}
    \end{center}
    \vspace{-0.1in}
    \caption{Inference efficiency study of our \model.}
    \label{fig:infer_time}
    \vspace{-0.2in}
\end{figure}

\noindent\textbf{Model Inference Efficiency}.
% In our further exploration, we aim to assess the inference speed and accuracy of our \model, by compare our model with state-of-the-art open-sourced LLMs, including baichuan-7B, vicuna-7B-v1.1, and vicuna-7B-v1.5. The comparison is conducted on the Arxiv and Cora COT instruction datasets using a single 40G Nvidia A100 for inference time measurement (second per response), as depicted in Figure~\ref{fig:infer_time}. We can notice that our graph LLM achieves superior model efficiency and prediction accuracy compared with other alternatives. It's important to note that in this comparison, achieving lower inference time does not necessarily indicate better performance. While the baichuan-7B model exhibits shorter inference time, it often produces brief incorrect or irrelevant answers due to its limited ability to handle complex multi-classification problems. On the other hand, the vicuna-7B-v1.1 and vicuna-7B-v1.5 models require a complex step-by-step reasoning process with longer inference time to generate better answers. In contrast, our model leverages a brief reasoning process but obtain more accurate prediction results, thereby enhancing the model inference efficiency.
In our exploration, we assess the inference speed and accuracy of our \model\ by comparing it with baichuan-7B, vicuna-7B-v1.1, and vicuna-7B-v1.5 LLMs. Using a single 40G Nvidia A100, we measure inference time (seconds per response) on the Arxiv and Cora COT instruction datasets, as shown in Figure~\ref{fig:infer_time}. Our graph LLM demonstrates superior efficiency and accuracy. Lower inference time doesn't necessarily mean better performance: baichuan-7B yields quick but often incorrect or irrelevant answers, while vicuna-7B-v1.1 and vicuna-7B-v1.5 require longer, complex reasoning steps for better answers. In contrast, our model achieves accurate predictions through a brief reasoning process, enhancing inference efficiency.

\vspace{-0.1in}
\subsection{Model Case Study (RQ5)}
\begin{table*}[h]
\centering
\caption{Comparison of prediction results between our \model\ and ChatGPT.}
\vspace{-0.1in}
    \label{tab:case_1}
\small
\begin{tabularx}{1\textwidth}{X X}
    \toprule
     % \multicolumn{2}{c}{\textbf{Title}}  \\
     
     % \textbf{Input} & \textbf{Output} \\
     % Input: & Output: \\
      \textbf{Title:} TiM-DNN: Ternary in-Memory accelerator for Deep Neural Networks. (with \textbf{102} first-order and second-order neighbors)\\
      
      \midrule
      \textbf{Ground-Truth Category:} cs.LG, Machine Learning\\
      \midrule
      \textbf{ChatGPT (Node Content Only):} 
      cs.AR, cs.AI, cs.SY, cs.ET, cs.NE. The paper presents a hardware accelerator for deep neural networks (DNNs) that is specifically designed to execute ternary DNNs, which represent weights and activations by signed ternary values. Therefore, the most likely category for this paper is cs.AR (Hardware Architecture), which focuses on the design and analysis of hardware and systems for AI and machine learning applications. The paper also involves the optimization of compute and storage requirements of DNNs, which falls under the category of cs.AI (Artificial Intelligence). Additionally, the paper proposes a programmable, in-memory accelerator, which is a type of embedded system, and therefore, it could be categorized under cs.SY (Systems and Control). Furthermore, the paper discusses the use of ternary values for DNNs, which is an emerging technique to optimize the compute and storage requirements of DNNs, and therefore, it could be categorized under cs.ET (Emerging Technologies). Lastly, the paper proposes a new bit-cell called Ternary Processing Cells (TPCs), which could be categorized under cs.NE (Neural and Evolutionary Computing). \\
      \midrule
    \textbf{ChatGPT (Node Content and Text-based Graph Structure):} 
    Based on the title "TiM-DNN: Ternary In-Memory Accelerator for Deep Neural Networks", the paper is likely to belong to the following arXiv CS sub-categories: 1. cs.AR (Hardware Architecture): The paper proposes a hardware accelerator, TiM-DNN, specifically designed for executing ternary deep neural networks. This falls under the domain of hardware architecture research. 2. cs.NE (Neural and Evolutionary Computing): The paper focuses on accelerating deep neural networks, which is a topic within the field of neural and evolutionary computing. 3. cs.CV (Computer Vision and Pattern Recognition): Deep neural networks are widely used in computer vision tasks, and the TiM-DNN accelerator could potentially be applied to accelerate computer vision algorithms. 4. cs.SY (Systems and Control): The TiM-DNN accelerator is a system-level solution for accelerating deep neural networks, so it could be categorized under systems and control research. 5. cs.DC (Distributed, Parallel, and Cluster Computing): While the paper does not explicitly mention distributed or parallel computing, the TiM-DNN accelerator is designed to optimize the compute and storage requirements of deep neural networks, which are often executed on distributed or parallel computing systems. \\
      \midrule
      \textbf{\model:} Based on the title and abstract of the paper, we can identify the following arXiv CS sub-categories that are most likely to be relevant: 
      \setlength\intextsep{0pt}
      \begin{wrapfigure}{l}{0.25\linewidth}
        \includegraphics[width=\linewidth]{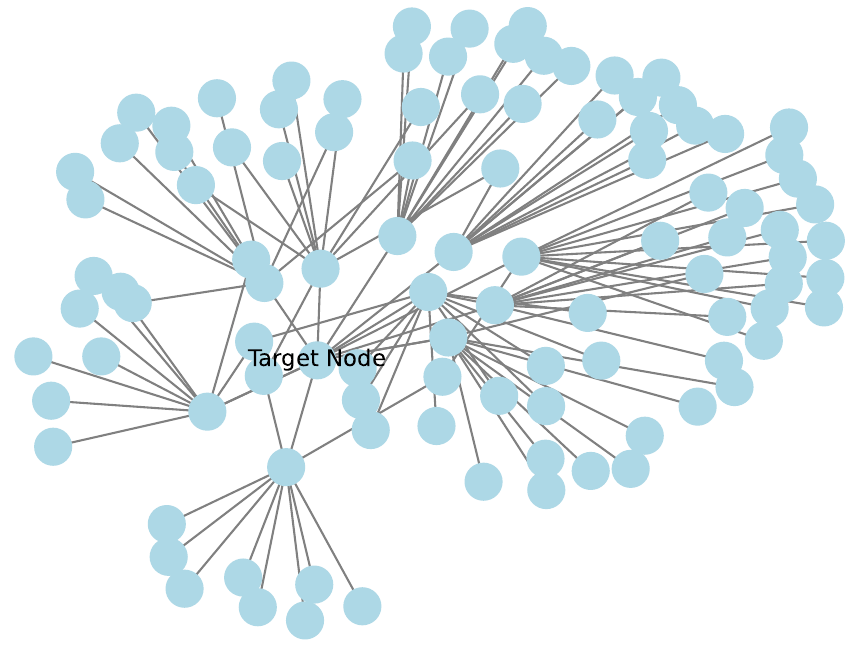}
    \end{wrapfigure}
    % \hspace{-0.3in}
    1. cs.LG (Machine Learning): This sub-category is relevant because the paper discusses the use of ternary DNNs for language modeling and image classification, which are both tasks that involve learning. 2. cs.AI (Artificial Intelligence): This sub-category is relevant because the paper proposes a new accelerator for DNNs, which is designed to optimize the compute and storage requirements of complex DNNs. 3. cs.NE (Numerical Analysis): This sub-category is relevant because the paper discusses the use of ternary DNNs, which involve signed ternary values, and the use of TiM tiles, which are specialized memory arrays that perform massively parallel signed vector-matrix multiplications on ternary values. 4. cs.SY (Systems and Control Theory): This sub-category is relevant because the paper discusses the implementation of TiM-DNN in 32nm technology using an architectural simulator calibrated with SPICE simulations and RTL synthesis. 5. cs.AR (Hardware Architecture): This sub-category is relevant because the paper proposes a new accelerator for DNNs, which is designed to be programmable and in-memory. \\

     \bottomrule
\end{tabularx}
\vspace{-0.15in}
\end{table*}

We conduct a detailed analysis of our model's performance in downstream graph learning tasks compared to traditional LLMs using different types of instructions. We evaluate ChatGPT and our \model\ using Arxiv data, with prompts based on node content, node content with text-based graph structure, and our designed graph instruction. The results, shown in Table~\ref{tab:case_1}, clearly demonstrate that despite its massive parameter count (over 200B), ChatGPT struggles to make accurate predictions solely based on node text information or node content with text-based graph structure. This challenge is particularly evident when dealing with papers that have cross-disciplinary characteristics, as seen in the example of research domains in machine learning and hardware architecture. In contrast, our \model\ consistently provides accurate predictions and reasonable explanations. This is because our \model\ incorporates a subgraph structure with 103 nodes, allowing it to extract rich structural information from neighboring nodes' citation relationships, leading to accurate predictions.

Furthermore, we believe that our approach of using graph tokens to represent the graph structure as input to the LLM is more efficient than the natural language solution. In the case of a 103-node subgraph, our \model\ only requires 750 tokens to be fed into the LLM, while the text-based method requires 4649 tokens. This significant reduction in token consumption translates to a substantial decrease in training and inference resource requirements.

\vspace{-0.05in}
\section{Related Work}
\label{sec:relate}

\noindent\textbf{Self-supervised Learning and Pre-training on Graphs}. To enhance the robustness of graph models, self-supervised learning (SSL) has been introduced as a powerful technique~\cite{liu2022graph,jing2021hdmi,hu2020gpt}. It allows GNNs to learn meaningful graph representations without heavily relying on labeled data. The core idea behind self-supervised learning in graph models is to design pretext tasks that leverage the graph's intrinsic properties to generate additional supervision signals~\cite{xia2022simgrace}. SSL-enhanced graph learning methods can be broadly classified into two main paradigms: contrastive SSL and generative SSL. In particular, i) \textbf{Contrastive SSL} focuses on learning representations by contrasting positive and negative samples. Notable methods in this domain include GraphCL~\cite{you2020graph} and GCA~\cite{zhu2021graph}. Recent advancements in contrastive SSL include automated contrastive augmentation (\ie, JOAO~\cite{you2021graph}, AdaGCL~\cite{jiang2023adaptive}), Hyperbolic-Euclidean dual space contrasting (\eg, DSGC~\cite{yang2022dual}), or community-aware contrastive learning (\eg, gCooL~\cite{li2022graph}). ii) \textbf{Generative SSL}, on the other hand, focuses on generating realistic samples that resemble the original graph structures. Recent advancements in this line include GraphMAE~\cite{hou2022graphmae,hou2023graphmae2} for feature masking, and S2GAE~\cite{tan2023s2gae}, AutoCF~\cite{xia2023automated} for reconstructing masked edges as SSL tasks. \\\vspace{-0.1in}

\noindent \textbf{Prompt-Tuning for Graph Neural Networks}. Recent efforts in enhancing the generalization capability of graph neural networks (GNNs) have focused on training GNN models in a self-supervised manner, followed by fine-tuning on specific downstream tasks using prompt-tuning techniques~\cite{zhang2023structure}. For example, GPPT~\cite{sun2022gppt} is a transfer learning paradigm, where GNNs are first pre-trained on masked edge prediction and then prompted with token pairs for downstream node classification. GraphPrompt~\cite{liu2023graphprompt} aims to handle downstream tasks by integrating pre-training and downstream tasks into a unified task template. Additionally, Sun~\etal~\cite{sun2023all} presents a unified prompt format, reformulates tasks to the graph level, and incorporates meta-learning techniques to improve multi-task performance in graph prompting. Despite these advances, these methods still require further fine-tuning that relies on supervision labels from downstream tasks to ensure accurate learning. In contrast, this work addresses this limitation by introducing a foundational graph model that tackles the more challenging task of zero-shot graph learning. By eliminating the need for label inputs from downstream tasks, this approach allows for a more general and flexible graph learning paradigm in real-world scenarios. \\\vspace{-0.1in}

\noindent \textbf{large Language Models}. In recent years, LLMs (\eg, ChatGPT~\cite{InstructGPT} and Claude~\cite{claude}) have gained widespread attention for their remarkable capabilities in various NLP tasks~\cite{LLM_Emergent,LLM_0shot}. Based on these unique capabilities of LLMs, many tuning-free prompting techniques have been explored to enhance their generative abilities, such as in-context learning~\cite{in_context} and Chain-of-Thought~\cite{cot, tot}. With the rise of open-source LLMs, such as Llama~\cite{llama,llama_2}, ChatGLM~\cite{GLM}, and Baichuan~\cite{baichuan}, technologies for aligning pre-trained LLMs to different specific tasks and human feedback have been proposed, making private LLMs in specific domains possible~\cite{instruct_tuning_2,Self_Instruct,RLAIF}.

While there have been successful attempts to align LLMs with visual information, such as multimodal LLMs~\cite{llava, minigpt4}, the alignment of LLMs with graph structures remains largely unexplored. This research addresses this gap by introducing a dual-stage graph instruction tuning paradigm that effectively aligns the language capacity of LLMs with graph learning. Previous studies~\cite{ GPT4Graph, potential_llm} have attempted to incorporate graph information into LLMs using natural language, but they have faced challenges in handling complex graph structures and achieving a deep understanding of graphs due to the limitations of relying solely on text-based prompts.

\vspace{-0.05in}
\section{Conclusion}
\label{sec:conclusion}

This work presents an effective and scalable graph large language model, aims at improving the generalization capabilities of graph models. The proposed framework, \model, injects graph domain-specific structural knowledge into the LLM through a dual-stage graph instruction tuning paradigm. By leveraging a simple yet effective graph-text alignment projector, we enable LLMs to comprehend and interpret the structural components of graphs. Extensive evaluations across different settings demonstrate the effectiveness of our method in both supervised and zero-shot graph learning scenarios. Furthermore, the model exhibits strong generalization abilities, allowing it to handle diverse downstream datasets and tasks without suffering from catastrophic forgetting. A potential avenue for future investigation is exploring pruning techniques to compress redundant or less important parameters of LLM, thereby reducing the overall model size while preserving its performance.

% A potential avenue for future investigation is exploring pruning techniques to compress redundant or less important parameters of LLM, thereby reducing the overall model size while preserving its performance.

% \textbf{Future Directions:} A potential avenue for future investigation could be:
% \textbf{i)} further exploring how to integrate more graph learning tasks into instruction datasets, achieving data-centric AI in graph learning.
% \textbf{ii)} exploring pruning techniques to compress redundant or less important parameters of LLM, thereby reducing the overall model size while preserving its performance.
% \textbf{iii)} scaling up parameters of our \model\ and investigating the emergent capabilities of LLMs in the graph-oriented domain.

\clearpage
\bibliographystyle{ACM-Reference-Format}
\balance
\bibliography{sample-base}

% \clearpage
% \input{appendix}

\end{document}